\documentclass[final,12pt]{elsarticle}

\usepackage{graphicx} 
\usepackage{amsmath}
\usepackage{amssymb}
\usepackage[dvipsnames]{xcolor}
\usepackage{epstopdf}
\usepackage{subcaption}
\usepackage{multirow}
\usepackage{placeins}
\usepackage{lineno}
\setcitestyle{square}

\DeclareMathOperator*{\argmin}{arg\,min}

\newcommand{\modif}[1]{{\color{black}#1}}

\journal{Renewable Energy}

\begin{document}

\begin{frontmatter}



\title{Active Learning-Based Optimization of Hydroelectric Turbine Startup to Minimize Fatigue Damage}


\author[inst1]{Vincent Mai}

\affiliation[inst1]{organization={Institut de recherche d'Hydro-Québec},
            addressline={1800 boul. Lionel Boulet}, 
            city={Varennes},
            postcode={J3X 1P7}, 
            state={Québec},
            country={Canada}}

\author[inst1]{Quang Hung Pham}
\author[inst1]{Arthur Favrel}
\author[inst1]{Jean-Philippe Gauthier}
\author[inst1]{Martin Gagnon}

\begin{abstract}

Hydro-generating units (HGUs) play a crucial role in integrating intermittent renewable energy sources into the power grid due to their flexible operational capabilities. This evolving role has led to an increase in transient events, such as startups, which impose significant stresses on turbines, leading to increased turbine fatigue and a reduced operational lifespan. Consequently, optimizing startup sequences to minimize stresses is vital for hydropower utilities. However, this task is challenging, as stress measurements on prototypes can be expensive and time-consuming. To tackle this challenge, we propose an innovative automated approach to optimize the startup parameters of HGUs with a limited budget of measured startup sequences. Our method combines active learning and black-box optimization techniques, utilizing virtual strain sensors and dynamic simulations of HGUs. This approach was tested in real-time during an on-site measurement campaign on an instrumented Francis turbine prototype. The results demonstrate that our algorithm successfully identified an optimal startup sequence using only seven measured sequences. 
It achieves a remarkable 42\% reduction in the maximum strain cycle amplitude compared to the standard startup sequence. This study paves the way for more efficient HGU startup optimization, potentially extending their operational lifespans.
\end{abstract}




\end{frontmatter}



\section{Introduction}

Hydropower accounts for more than 50$\%$ of renewable electricity production worldwide. In addition, as utilities incorporate an increasing share of intermittent renewable energy sources, such as wind and solar, into Electrical Power Systems (EPSs), hydropower is expected to play a pivotal role in the energy transition. Hydro-generating units (HGUs) are characterized by their rapid startup and shutdown sequences, as well as their ability to adjust power output quickly. They can operate effectively from no-load to full-load conditions and are capable of responding swiftly to fluctuations in the power output of intermittent renewable sources and variations in electricity grid frequency. For these reasons, they provide a growing array of ancillary services to EPS \citep{Vagnoni2024}. This evolving role translates into more frequent transients, including startup and shutdown operations, as well as functioning under off-design conditions \modif{ having significant influence on the turbine runner life \citep{ChiragTrivedi_Michel_2013}}. Such conditions increase runner fatigue and the risk of early failure due to adverse hydrodynamic phenomena, which impose significant stresses on the runner \citep{Liu_Luo_Wang_2016}. Consequently, the study of these conditions has garnered \modif{more }attention over the past decade within both academic and industrial communities, through experimental campaigns on reduced-scale models \citep{Duparchy2015}\modif{, correlation between reduced-scale models and prototype machines \citep{Favrel2020}, modeling of the fatigue loading from prototype machines measurements \citep{Pham_Gagnon_Antoni_Tahan_Monette_2021},} as well as through Computational Fluid Dynamics (CFD) and Finite Element Analysis (FEA) numerical simulations \modif{to capture the blade fatigue loading \citep{Nicolle2012}. Furthermore,  the stochastic nature of those events was also studied using both prototype machine measurements \citep{Gagnon2012} and CFD-FEA numerical simulations \citep{Morissette2019}.}

Among these conditions, HGU startup sequences induce substantial stress on the runner \citep{Gagnon_Tahan_Bocher_Thibault_2010}. This poses a significant challenge for hydropower utilities, particularly in light of the anticipated increase in starting cycle frequency \citep{Savin_Badina_2020}. Over the lifetime of the turbine, the stress cycles experienced by the runner contribute to fatigue, leading to crack propagation \citep{Gummer_Etter_2008}. The risk of blade cracking increases the frequency of the required inspections, which results in significant financial costs, reduced flexibility, and a loss of power output for hydropower utilities \citep{Nilsson_Sjelvgren_1997}. Therefore, mitigating turbine stresses during startup is crucial to enhance the reliability and longevity of hydroelectric plants in the evolving energy landscape as shown in \citep{Gagnon2018}.

Stress levels experienced by hydroturbines during startup are closely linked to the trajectory in terms of guide vane opening and rotational speed \modif{as highlighted in both \citep{Cepa_Alic_Dolenc_2018} and \citep{Mukai_2022}}. The guide vane opening sequence \modif{is controlled by the hydraulic turbine governing systems. The governing system} is parameterized and ensures that the turbine reaches synchronous speed while respecting specific operational requirements \modif{ \citep{IEC61362_2024}}. Optimization of startup sequence parametrization \modif{has been addressed by several researchers with different approaches}. \modif{For example, Seydoux et al. \citep{Seydoux2024} and Muser et al. \citep{Muser2025} proposed optimization framework based on Voronoi cell tessellation and deep-learning algorithms, respectively, applied to experimental reduced-scale model data, while Unterluggauer et al. \citep{UNTERLUGGAUER2020} compared the impact of different startup scenario on turbine life expectancy through on-site campaign on a full-scale machine. From a numerical point of view, Schmid et al. \citep{Schmid2022} proposed an optimization framework combining a runner damage hillchart derived from CFD-FEA simulations with 1D hydraulic transient simulations}.

Model testing allows for cost reductions and exploration of a wide variety of startup scenarios without the access and time constraints encountered in prototype tests. However, the transposition of turbine mechanical responses during startup from model to full-scale remains a significant challenge, and no rigorous methodology has been established to date. In contrast, prototype measurements provide insights into the real mechanical behavior of full-scale turbines. However, these measurements are challenging and costly for hydroelectric operators due to time constraints and the harsh environment in which hydroelectric turbines operate, which often leads to the rapid degradation of strain gauges. Therefore, startup optimization must rely on the limited data acquired during short, dedicated measurement campaigns.

\modif{Gagnon et al., in 2014 \citep{Gagnon_Jobidon_Lawrence_Larouche_2014} and 2016 \citep{Gagnon2016}}, presented the results of such campaigns, highlighting the effects of startup sequences on blade stresses. Based on their results, they defined optimal startup scenarios in terms of stress levels. However, these scenarios are mainly derived through trial-and-error processes and expert experience, leaving the potential for further optimization untapped. While optimization may be performed post-campaign using a virtual sensor approach \citep{Gagnon_Pham_Mai_Favrel_2023}, the results cannot be validated without a subsequent experimental campaign. 

\modif{Given the limited strategies found to optimize startup transients in the literature presented in this study,} we propose a novel methodology that allows for the optimization of startup trajectory parameters during a measurement campaign, ultimately contributing to extending the life of hydroturbines. By combining an active learning framework with virtual sensors and black-box optimization, the active learning algorithm sequentially generates various startup parameters to test on the turbine. At each step, new data are used to refine the model and optimization process. This approach was successfully tested during an experimental campaign conducted on a full-scale Francis turbine instrumented with several strain gauge rosettes.  

This article is structured as follows. In Section~\ref{sec:problem_description}, the problem is formulated in details, from HGU dynamics and strain generation to the resulting active optimization problem. Section~\ref{sec:methodology} presents the proposed methodology, which combines an HGU dynamics simulator, virtual strain gauges, black-box optimization and the active learning approach. This method was applied to a real HGU during a measurement campaign, and the results are provided and discussed in Section~\ref{sec:experimental_results}. \modif{To facilitate the reading of the paper, a mathematical notation table is provided in appendix \ref{app:notation_table}.}


\section{Problem description}
\label{sec:problem_description}

When disconnected from the power grid, which is the case during startup, HGU dynamics are determined by the turbine characteristics. For a given hydraulic head, the net mechanical torque applied on the turbine by the flow is determined by the rotational speed $\omega(t)$ and the guide vane opening $o(t)$. In this paper, $\omega$ is normalized by the HGU's synchronous speed $\omega_S$, and $o$ by the maximum vane opening. For any given opening, there exists a corresponding speed where the net torque is zero and the unit remains at steady-state equilibrium. Conversely, increasing and decreasing the guide vane opening from equilibrium at any speed will respectively result in acceleration and deceleration. Control of the rotational speed can thus be achieved by varying the guide vane opening.

During startup, the unit is brought up to synchronous speed $\omega_S$ by opening the guide vanes according to a sequence determined by the speed governor. The speed governor includes an electronic control system and a hydraulic valve circuit leading to the servomotors, which are connected to the guide vanes. The electronics generate the guide vane setpoint $u(t)$, which is used in combination with feedback on the rotational speed $\omega(t)$ and guide vane opening $o(t)$ to create the command signal controlling the hydraulics. The governor's configuration and the inertia of the moving parts result in the guide vane opening $o(t)$ not reacting instantly to setpoint changes. 

\begin{figure}[h]
    \centering
    \includegraphics[width=0.9\textwidth]{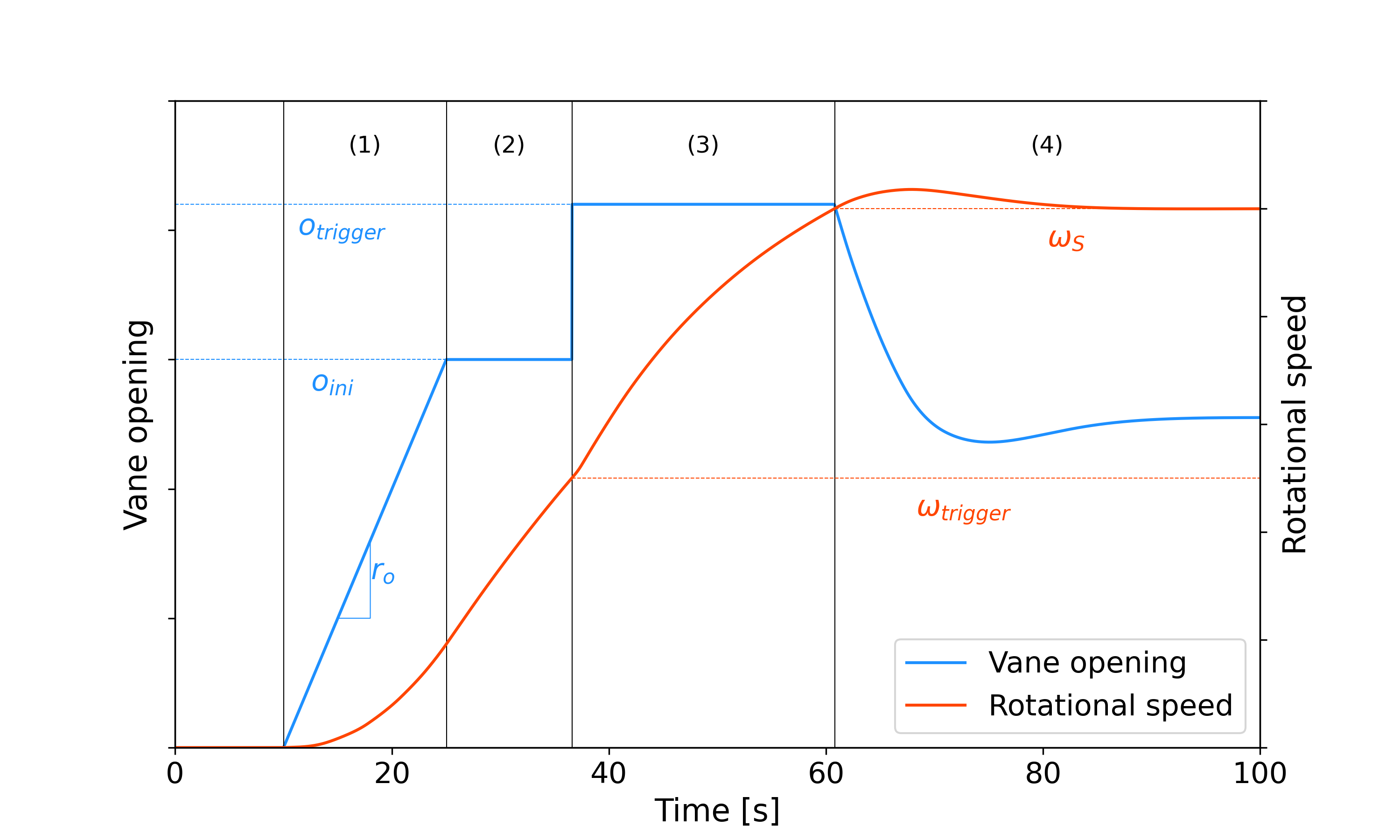}
    \caption{Startup sequence parameters and phases}
    \label{fig:sequence_demarrage}
\end{figure}

The setpoint signal $u(t)$ is created dynamically; its exact shape is determined by four tunable parameters, included in set $\theta$:
\begin{equation}
    \label{eq:definition_theta}
    \theta = \left\{r_{\mathrm{o}}, o_{\mathrm{ini}}, \omega_{\mathrm{trigger}}, o_{\mathrm{trigger}} \right\}
\end{equation}
with $r_{\mathrm{o}}$, the opening rate, $o_{\mathrm{ini}}$, the initial opening, $\omega_{\mathrm{trigger}}$, the trigger rotational speed, and $o_{\mathrm{trigger}}$, the trigger opening.  These parameters $\theta$ control the four phases of the startup process, which are illustrated in Figure~\ref{fig:sequence_demarrage}:
\begin{enumerate}
    \item Ramp-up: Starting at zero, $u(t)$ increases at the rate $r_{\mathrm{o}}$ until it reaches the value $o_{\mathrm{ini}}$.
    \item First plateau: $u(t)$ remains at the constant value $o_{\mathrm{ini}}$ until $\omega(t)$ reaches the trigger speed $\omega_{\mathrm{trigger}}$.
    \item Second plateau: $u(t)$ steps up instantaneously to $o_{\mathrm{trigger}}$ and remains at this constant value until $\omega(t)$ reaches $\omega_S$. 
    \item Feedback control: The PID controller becomes effective and automatically manages $u(t)$ to bring back $\omega(t)$ at the value $\omega_S$ until it is close and stable enough for synchronization.
\end{enumerate}

\begin{figure}[h]
    \centering
    \includegraphics[width=\textwidth]{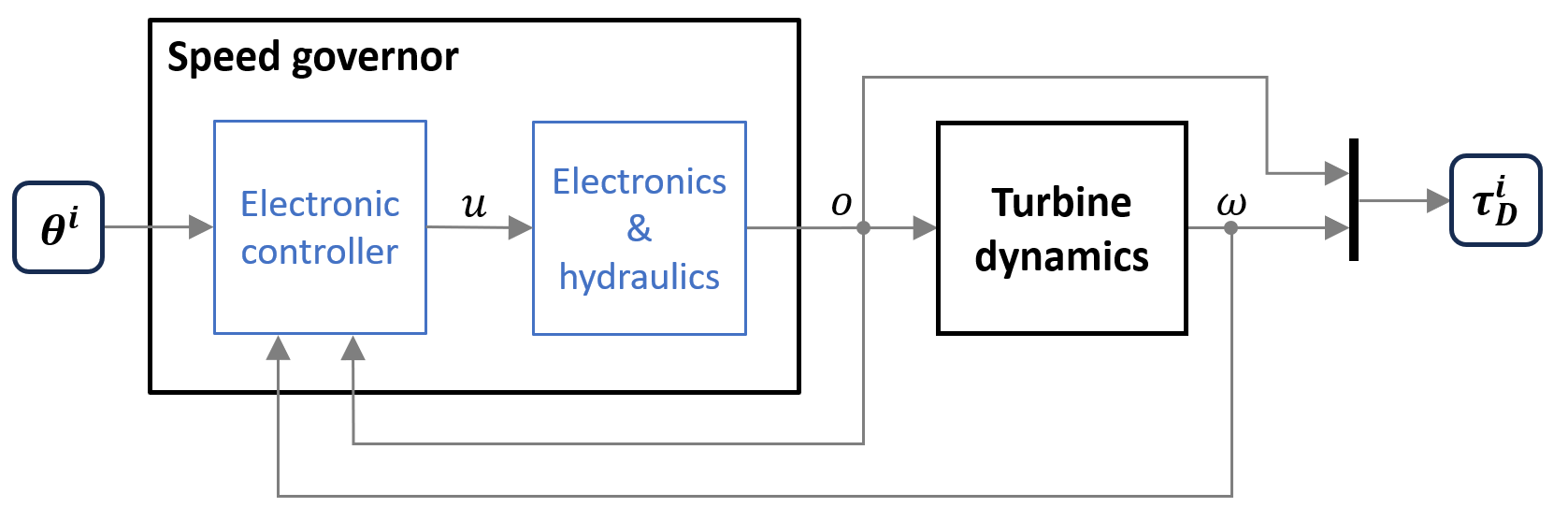}
    \caption{HGU dynamics block diagram}
    \label{fig:schema_dynamique_GTA}
\end{figure}

Under the hypothesis that HGU dynamics are deterministic, two startups with identical parameter sets $\theta^i$ lead to the same dynamic trajectory $\tau_D^i$. This trajectory is discretized in time at frequency $f_D$, and defined as:
\begin{equation}
    \tau_D^i \doteq (\omega^i_0, o^i_0, \omega^i_1, o^i_1, ..., \omega^i_{n^i_{D,\mathrm{st}}}, o^i_{n^i_{D,\mathrm{st}}})
\end{equation}
where $n^i_{D,\mathrm{st}}$ is the time step at which the turbine is ready for synchronization, corresponding to a startup time of $t^i_{\mathrm{st}} = n^i_{D,\mathrm{st}}/f_D$. Figure~\ref{fig:schema_dynamique_GTA} shows a graphical representation of the HGU dynamics as a high-level block diagram. 
\FloatBarrier

At time $t$, the strain $s(t)$ at a given location of a turbine blade can be measured by strain gauges, usually in microstrains ($\mu$S), at frequency $f_M$. During a startup process, the measured trajectory can be captured as $\tau_M^i$, defined as the combination of the dynamic trajectory $\tau_D^i$ and the strain measurements $s^i_n$:
\begin{equation}
    \tau_M^i \doteq \left(\omega^i_0, o^i_0, s^i_0, \omega^i_1, o^i_1, s^i_1, ..., \omega^i_{n^i_{M, \mathrm{st}}}, o^i_{n^i_{M, \mathrm{st}}}, s^i_{n^i_{M, \mathrm{st}}} \right)   
    \label{eq:trajectory}
\end{equation}
where the last time step $n^i_{M, \mathrm{st}} = t^i_\mathrm{st} f_M$.

On turbine blades, repeated high-amplitude strain cycles lead to fatigue and potential cracks \citep{Gagnon_Thibault_2015}. The greater the cycle amplitude, the greater the accumulated damage to the blade over time. Therefore, in fatigue analysis, the largest amplitude strain/stress cycle, meaning the difference between the highest and lowest strain values over the complete start process, has the most impact on fatigue. As a result, the objective of minimizing fatigue during startup can be approximated by minimizing the amplitude of the largest strain cycle. Formally, this can be expressed as minimizing the following loss over a given trajectory:

\begin{equation}
    \mathcal{L}(\tau_M^i) = \max_{n \in \nu_M^i}(s^i_n) - \min_{n \in \nu_M^i}(s^i_n)
\end{equation}
where $\nu_M^i \doteq (0, ..., n^i_{M,\mathrm{st}})$.

Due to the complexity of fluid-structure interaction phenomena occurring during turbine startup, strain dynamics can be considered a high frequency stochastic process, where $\tau_M^i$ is conditioned on dynamic trajectory $\tau^i_D$. As $\tau^i_D$ is characterized by startup parameters $\theta^i$, it is possible to define the conditional probability distribution $p_s(\tau^i_M | \theta^i)$ and the corresponding loss distribution $p_l(\mathcal{L}(\tau_M^i)| \theta^i)$.

Optimizing startups can be formulated as a constrained optimization problem. The optimal solution is the control parameters $\theta^*$ which minimize the expected maximal strain amplitude of the largest cycle over the entire trajectory.

\begin{equation}
\label{eq:opt_loss}
    \theta^* = \argmin_\theta \mathbb{E}_{p_l}\left[ \left. \mathcal{L}(\tau_M) \right|  \theta \right]
\end{equation}
under a time constraint related to reaching synchronous speed $\omega_S$ within the operational constraints: 
\begin{equation}
\label{eq:opt_const}
t^*_{\mathrm{st}} < \bar{t}_\mathrm{st}
\end{equation}

In the context of a measurement campaign for a specific hydro-electric turbine, access to an instrumented turbine is assumed, with no prior data. However, there is a budget of $N$ measured trajectories $\tau_M^i$ for which parameters $\theta^i$ can be chosen sequentially.
This is an active learning problem where the goal is to sequentially generate parameters $\theta^1$ to $\theta^N$ such that the expected loss of $\tau_M^N$ is close to the optimal expected loss of $\tau_M^*$. It is then possible to measure the strains over $\tau_M^N$ to evaluate the optimized parameters $\theta^N$.


\section{Methodology}
\label{sec:methodology}

We addressed the active learning problem defined in the previous section by introducing two models: an HGU simulator and strain virtual sensors. These are then combined in a black-box optimization loop, which itself is integrated into an active learning loop. 

\subsection{Startup trajectory simulation}

To accurately simulate the trajectory $\tau_D$, a simplified HGU simulator was developed. It comprises two parts:
\begin{enumerate}
    \item A speed governor model, with the adjustable parameters $\theta$, the rotational speed $\omega(t)$ as input, and the guide vane opening $o(t)$ as output;
    \item A turbine dynamics model, with the guide vane opening $o(t)$ as input and the rotational speed $\omega(t)$ as output.
\end{enumerate}

The speed governor model includes a PID controller with feedback loops on speed $\omega(t)$ and opening $o(t)$ and  reconstruction of the setpoint $u(t)$ based on the parameters $\theta$. The model accurately represents the transfer function between the setpoint $u(t)$ and the guide vane opening $o(t)$. 

Turbine dynamics can be addressed using the SIMSEN software \citep{Nicolet_2007}. A SIMSEN model of the HGU includes the upstream and downstream reservoirs, the water piping system, and the turbine. It can be used to simulate the rotational speed $\omega(t)$ for a predefined guide vane opening $o(t)$ time series. A typical simulation of a single HGU startup takes between 30 seconds and 1 minute. This is problematic for an active optimization framework, where hundreds of simulations must be performed in a limited time. Moreover, in practice, the guide vane opening $o(t)$ is not known beforehand; instead, it is calculated as part of the speed governor model, using the rotational speed $\omega(t)$. To properly address this bidirectional coupling, a co-simulation approach would have to be used in order to ensure communication between the SIMSEN model and speed governor model at each time step, further increasing computational time.

\begin{figure}[h]
    \centering
    \includegraphics[width=0.7\textwidth]{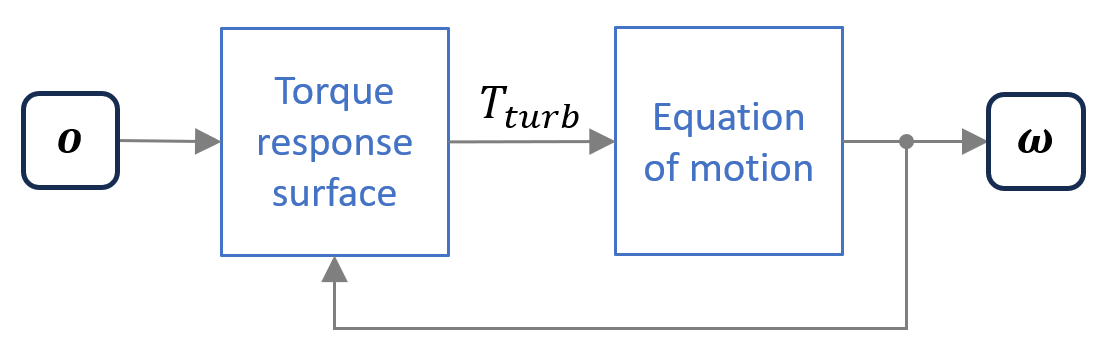}
    \caption{Turbine dynamics model}
    \label{fig:modele_turbine}
\end{figure}

To overcome these limitations, a quasi-static surrogate model was used instead. In the SIMSEN model of the HGU, the gross hydraulic head was set at its nominal value since significant deviations are not expected. Steady-state simulations covering a uniform grid over a selected range in the parametric space $(\omega,o)$ were carried out and the simulated torque was recorded for each run. A bilinear interpolant was fitted to the resulting data points, creating a response surface that can be evaluated very quickly for any combination of guide vane opening and rotational speed. The rotational speed $\omega(t)$ can be obtained by solving the equation of rotational motion of the HGU:
\begin{equation}
\label{eq:mouvement_GTA}
    J \frac{\mathrm{d}\omega}{\mathrm{d}t} = T_\mathrm{turb}
\end{equation}
where $J$ is the total inertia of the rotating components (including the generator, shaft, and turbine) and $T_{turb}$ is the turbine torque. No resistive electromagnetic torque from the generator is applied since the unit is off grid during the startup sequence. Also, bearing friction is neglected as it remains very low relative to $T_{turb}$. Figure~\ref{fig:modele_turbine} shows the block diagram of the turbine dynamics.

The HGU simulator is built by combining Equation~\ref{eq:mouvement_GTA} with the underlying equations of the speed governor model described above, resulting in a system of $N_{eq}$ nonlinear ordinary differential equations with adjustable parameters $\theta$:
\begin{equation}
\label{eq:systeme_eq_GTA}
    \frac{\mathrm{d}\mathbf{q}}{\mathrm{d}t} = \mathbf{F}(t,\mathbf{q},\theta)
\end{equation}
with the unknowns' vector $\mathbf{q}(t) = [\omega(t),o(t),u(t),q_4(t),q_5(t),\ldots,q_{N_{eq}}(t)]^T$, where $q_4(t)$ through $q_{N_{eq}}(t)$ are internal variables of the speed governor model that are not relevant for the optimization problem. 

\begin{figure}[h]
    \centering
    \includegraphics[width=\textwidth]{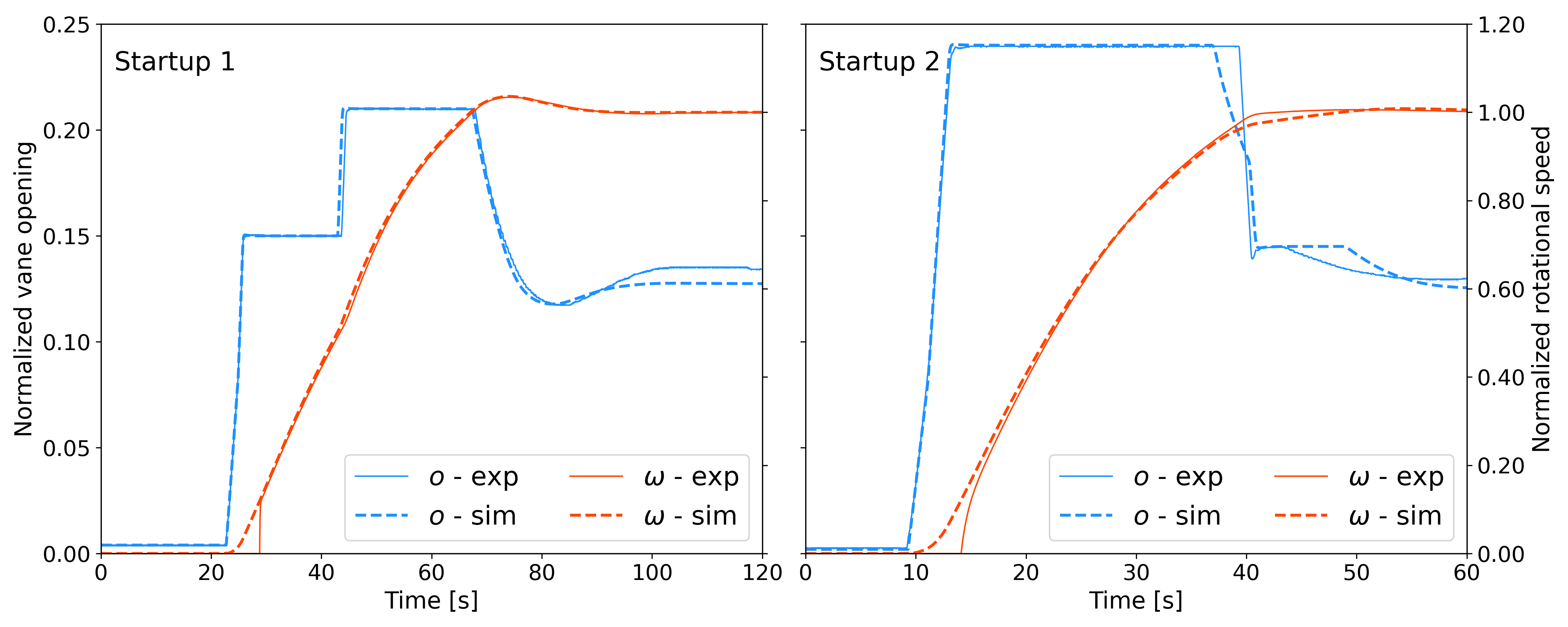}
    \captionsetup{justification=centering}
    \caption{Comparison of simulated startup sequences with experimental data}
    \label{fig:validation_simulateur}
\end{figure}

A simulation is carried out by solving the system using a classical fourth-order Runge-Kutta method, producing a dynamic trajectory $\tau_D^i$ with a startup time $t^i_{st}$ given a set of startup parameters $\theta^i$. The simulation is stopped when the rotational speed and acceleration are respectively equal to $\omega_S$ and zero within a certain tolerance, or when the simulated time reaches twice the constraint $T_{st}$, whichever occurs first. The simulator was validated against two real HGU startups with different parameter sets $\theta$. Figure~\ref{fig:validation_simulateur} shows that there is good agreement between the measured data and simulation data for both cases. The discontinuities observed in the experimental speed signals are due to the sensors inability to measure low rotational speeds. This approach does not require co-simulation, and simulating a startup takes approximately 1 second.

\FloatBarrier

\subsection{Strain envelope}

To optimize the quantity expressed in Equation~(\ref{eq:opt_loss}), it is possible to consider only the envelope of the strain, i.e., the upper $s^u_n$ and lower $s^l_n$ values over a certain time window $w$ defined as a number of steps. The envelope is defined as:
\begin{align}
    s^u_n & \doteq \max_{k \in [\lceil-w/2\rceil, \lceil w/2 \rceil]} (s_{n-k})     \label{eq:def_envelope_u}\\
    s^l_n & \doteq \min_{k \in [\lceil-w/2\rceil, \lceil w/2 \rceil]} (s_{n-k})
    \label{eq:def_envelope_l}
\end{align}

\begin{figure}
    \centering
    \includegraphics[width=0.9\linewidth]{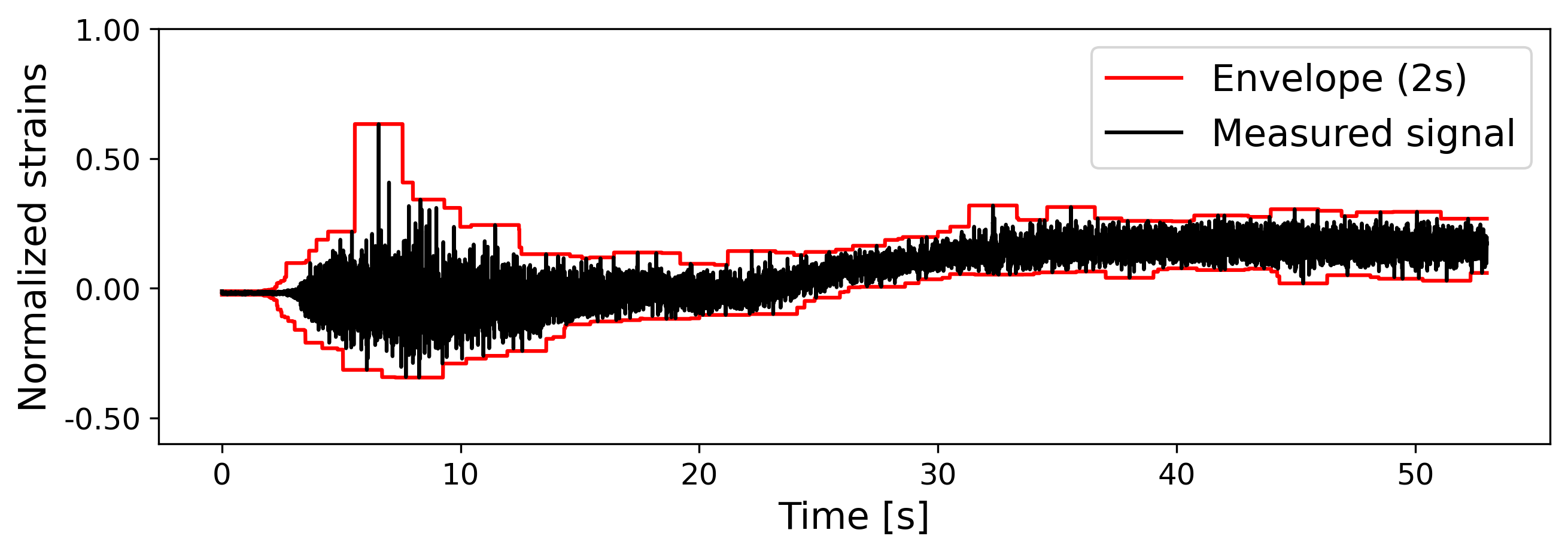}
    \caption{Example of startup measured strains and the corresponding resampled envelope}
    \label{fig:example_env}
\end{figure}

This transformation is useful because, compared to the strain signal $s(t)$ and its measurements $s_n$, the envelopes $s^u_n$ and $s^l_n$ evolve on significantly larger time scales. Once the envelopes are computed, they can be down sampled without a significant loss of information at a lower frequency $f_e$, leading to enveloped trajectories $\tau^i_e$:

\begin{equation}
\tau^i_e \doteq \left(\omega^i_0, o^i_0, s^{u,i}_0, s^{l,i}_0, \omega^i_1, o^i_1,  s^{u,i}_1, s^{l,i}_1, ..., \omega^i_{n^i_{e,\mathrm{st}}}, o^i_{n^i_{e,\mathrm{st}}}, s^{u,i}_{n^i_{e,\mathrm{st}}}, s^{l,i}_{n^i_{e,\mathrm{st}}}\right) 
\end{equation}
where $n^i_{e,\mathrm{st}} = t^i_{\mathrm{st}} f_e$. Figure~\ref{fig:example_env} shows an example of measured strains during a startup and the envelope computed over a window $w$ corresponding to 2 s.

The envelope's distribution can be modeled conditioned on the current rotational speed $\omega_n$ and guide vane opening $o_n$. At time step $n$, the conditional probability distribution functions $b^u$ and $b^l$ are defined as:
\begin{align}
    s^u_n &\sim b^u(\cdot | \omega_n, o_n) \\
    s^l_n &\sim b^l(\cdot | \omega_n, o_n) 
\end{align}

An important property of the envelope bounds is that, given $\nu_e = 0, 1, ..., n^i_{e,\mathrm{st}}$, the following is true:
\begin{align}
    \max_{n \in \nu_M}(s_n) & = \max_{n \in \nu_e}(s^u_n) \\
    \min_{n \in \nu_M}(s_n) & = \min_{n \in \nu_e}(s^l_n)
\end{align}
and thus $\mathcal{L}(\tau_M^i) = \max_{n \in \nu_e^i}(s^u_n) - \min_{n \in \nu_e^i}(s^l_n)$. This allows us to rewrite the optimization objective of Equation~(\ref{eq:opt_loss}) as a function of $b^u$ and $b^l$:

\begin{equation}
\label{eq:opt_loss_env}
    \theta^* = \argmin_\theta \mathbb{E}_{b^u, b^l}\left[ \left. \max_{n \in \nu_e}(s^u_n) - \min_{n \in \nu_e}(s^l_n) \right|  \theta \right]
\end{equation}
under the same time constraint as described in Equation~(\ref{eq:opt_const}).

\subsection{Strain modeling with virtual sensors}
\label{sec:virtual_sensor}
Evaluating the objective in Equation~(\ref{eq:opt_loss_env}), requires an estimation of $b^u$ and $b^l$. Given strain measurement data, it is possible to build a model of these distributions using the virtual sensor approach \citep{Gagnon_Pham_Mai_Favrel_2023}. 
$b^u(\cdot|\omega_n, o_n)$ and $b^l(\cdot|\omega_n, o_n)$ are \modif{modeled as} conditional normal distributions, parameterized by mean $\mu^u_n$ and $\sigma^u_n$ (respectively,  $\mu^l_n$ and $\sigma^l_n$). \modif{The normal distribution was chosen because it is simple and generally a good approximation. This assumption was not validated in the current study.} These parameters are estimated using the deep learning model $m_\phi(\omega_n, o_n) = \left(\hat{\mu}^u_n, \hat{\sigma}^u_n,\hat{\mu}^l_n, \hat{\sigma}^l_n \right)$ trained on a dataset $\mathcal{D}$ of enveloped trajectories $\{\tau_e\}$.

Model $m_\phi$ is parameterized as a feed-forward neural network composed of one input layer of dimension 2 ($\omega_n, o_n)$, two hidden layers of dimension 32, and one output layer of dimension 4 (two $\mu$ values and two $\sigma$ values). The hidden layers are normalized using batch normalization \citep{Ioffe_Szegedy_2015}. \modif{As demonstrated in \citep{Gagnon_Pham_Mai_Favrel_2023}, the strain envelope can be effectively predicted given current valve opening and rotational speed values for a given head.} 
Because the trajectories are time series, it could seem intuitive to use recurrent neural networks such as LSTM or GRU and condition the prediction over the past trajectory $\tau_D$. However, such architectures do not produce better results experimentally, hinting that, during startups, \modif{the current values} contain the necessary information to predict the strain envelope.

The model is trained on a dataset $\mathcal{D}$ of enveloped trajectories $\{\tau^i_e\}$, obtained from available measured trajectories $\{\tau_M^i\}$. During training, all time steps of all trajectories are randomly mixed to reduce the correlation between the various input-label samples $(\omega_n, o_n; s^u_n, s^l_n)$. 

Since $b^u$ and $b^l$ are modeled as conditional normal distributions, $m_\phi$ was trained using the $\beta$-negative log-likelihood loss function \citep{Seitzer_2022}, which was shown to robustly estimate heteroscedastic aleatoric uncertainty. Note that predicting $\mu^u_n$ and $\sigma^u_n$ is equivalent to estimating the aleatoric uncertainty of $s^u_n$ (and respectively for $s^l_n$). Model $m_\phi$ was trained over three epochs, using the Adam optimizer with a learning rate of 0.001 and a mini-batch size of 32. 

The active learning framework requires an estimation of the epistemic uncertainty of the model's prediction. This is achieved by training an ensemble of five neural network models, initialized randomly and independently and trained with data presented in a different order. This allows us to approximate the distribution of model parameters that would fit the data. 
The ensemble's prediction is computed as the mean of the networks' predictions, and the epistemic uncertainty $\sigma_{\mathrm{ep}, n}^{\{u,l\}}$ as their sampled standard deviation.

\subsection{Black-box optimization}
\label{sec:black-box_simple}
Given a virtual sensor $m_\phi(\omega_n, o_n)$, the objective of Equation~(\ref{eq:opt_loss_env}) under the time constraint of Equation~(\ref{eq:opt_const}) can be addressed using black-box optimization. We use the NOMAD 4 \citep{Audet_LeDigabel_Montplaisir_Tribes_2022} with a budget of $N_I$ calls to the black-box. 

In this case, the black box consists of the dynamics simulator, the trained virtual sensor model, and a cost evaluator. For a given $\theta^i$, the simulator provides the dynamic trajectory $\tau_D^i$. Then, each point $(\omega^i_n, o^i_n)$ of $\tau_D^i$ is passed through $m_\phi$, producing a simulated trajectory $\tau^i_S$ with frequency $f_S = f_D$:
\begin{equation}
    \tau^i_S \doteq \left(\omega^i_0, o^i_0, \hat{\mu}^{u,i}_0,  \hat{\sigma}^{u,i}_0, \hat{\mu}^{l,i}_0,  \hat{\sigma}^{l,i}_0, ..., \omega^i_{n^i_{S,\mathrm{st}}}, o^i_{n^i_{S,\mathrm{st}}}, \hat{\mu}^{u,i}_{n^i_{S,\mathrm{st}}},  \hat{\sigma}^{u,i}_{n^i_{S,\mathrm{st}}}, \hat{\mu}^{l,i}_{n^i_{S,\mathrm{st}}},  \hat{\sigma}^{l,i}_{n^i_{S,\mathrm{st}}}  \right)  
\end{equation}
where $n^i_{S,\mathrm{st}} = t^i_{\mathrm{st}} f_S$.
This allows us to compute a cost function $c(\theta^i)$ from $\tau^i_S$. More precisely, the cost is composed of two elements: $c(\theta^i) = \alpha_d c_s(\theta^i) +  c_c(\theta^i)$, where $c_s(\theta^i)$ is the strain cost, $c_c(\theta^i)$ is the time constraint cost, and $\alpha_d$ is a normalization factor. The strain cost is defined as follows:

\begin{equation}
    \label{eq:strain_cost}
    c_s(\theta^i) = \max_{n \in \nu_S} \left(\hat{\mu}^{u,i}_n + \hat{\sigma}^{u,i}_n \right) - \min_{n \in \nu_S} \left(\hat{\mu}^{l,i}_n + \hat{\sigma}^{l,i}_n \right)
\end{equation}
with $\nu_S = 0, 1, ..., n^i_{S,\mathrm{st}}$.
The aleatoric standard deviation $\hat{\sigma}$ is added to the mean $\hat{\mu}$ on both sides to reflect the fact that the objective in Equation~(\ref{eq:opt_loss_env}) is to minimize the difference of the expected \textit{extrema} of random variables $s^u$ and  $s^l$ over the trajectory. As $m_\phi$ outputs an estimation of the parameters of their distributions $b^u$ and $b^l$, modeled as normal distributions, we make the hypothesis that the standard deviation represents the likely extrema around the mean of the envelope when sampled around the maximum values. Note that, when using an ensemble of models, the  $\hat{\mu}^{\{u,l\},i}_n + \hat{\sigma}^{\{u,l\},i}_n $ sums used in Equation~(\ref{eq:strain_cost}) are computed as the means of the individual sums calculated for each network in the ensemble.

The constraint cost is related to the compliance of the time constraint expressed in Equation~(\ref{eq:opt_const}). This cost is expressed three separate cases, as:

\begin{equation}
    \label{eq:time_cost}
    c_c(\theta^i) = 
    \begin{cases}
        0 & \text{if } t^i_{\mathrm{st}} < 0.5 \bar{t}_\mathrm{st}\\
        0.05 (t^i_{\mathrm{st}} - 0.5 \bar{t}_\mathrm{st})/(0.5 \bar{t}_\mathrm{st}) & \text{if }  0.5 \bar{t}_\mathrm{st} \leq t^i_{\mathrm{st}} < \bar{t}_\mathrm{st} \\
        1  + \left( t^i_{\mathrm{st}} - \bar{t}_\mathrm{st} \right)/ (0.2 \bar{t}_\mathrm{st})  & \text{if } t^i_{\mathrm{st}} \geq \bar{t}_\mathrm{st}  \\
    \end{cases}
\end{equation}

In the second case, the time constraint is respected. However, $c_c$ signals to NOMAD that, given equal strain amplitude, a shorter trajectory is preferred. 
The last case, representing a violation of the time constraint, is set as an indicator of 1, plus a penalty that increases as the scale of the constraint violation increases to guide NOMAD towards shorter trajectories. If $\omega$ has not stabilized at $\omega_S$ after $2 \bar{t}_\mathrm{st}$, the simulation is stopped and $t^i_\mathrm{st}$ is set to $2 \bar{t}_\mathrm{st}$.

\begin{figure}
    \centering
    \includegraphics[width=0.6\linewidth]{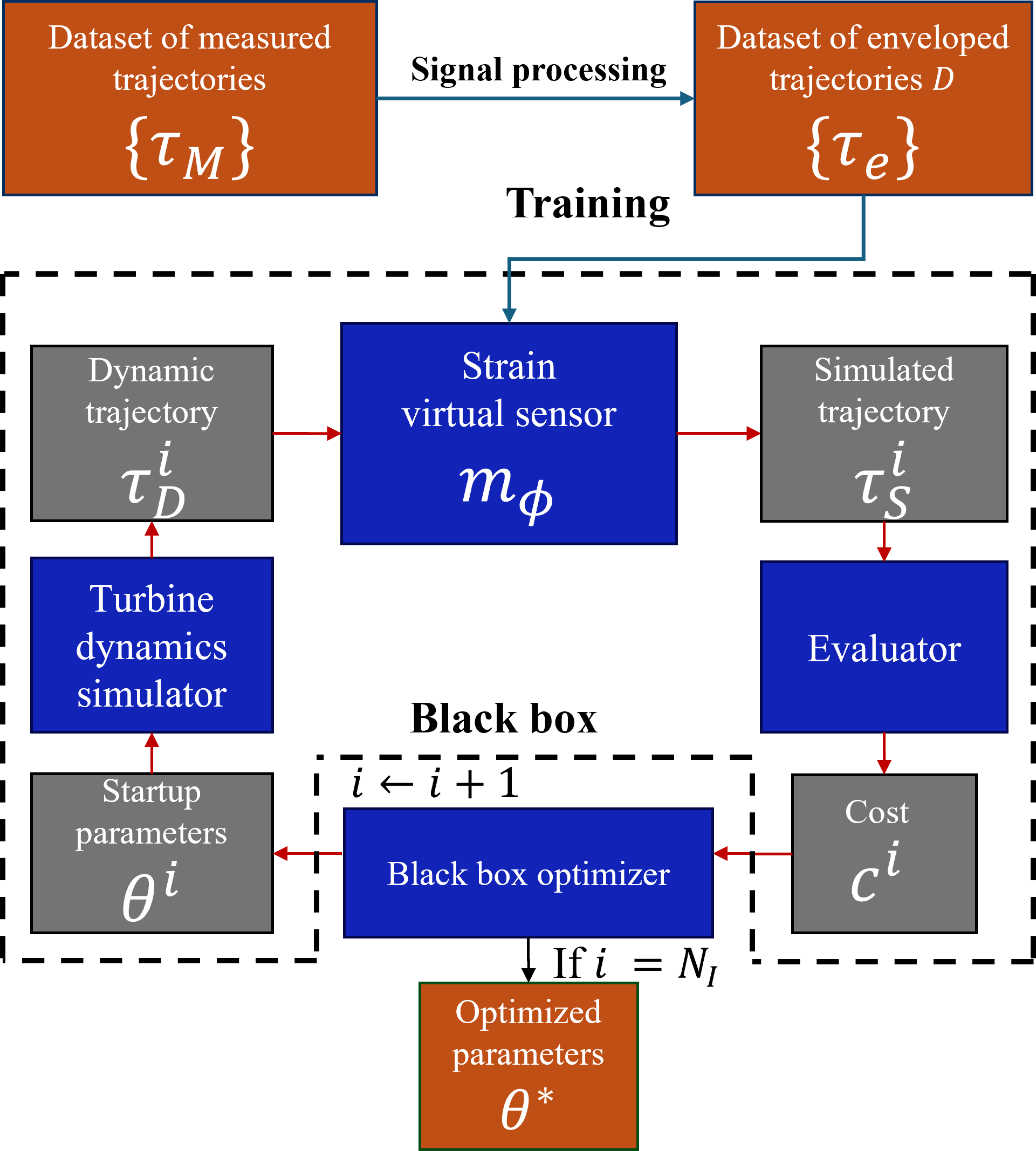}
    \caption{\modif{Different components of the black box given measured trajectories dataset $\mathcal{D}$}}
    \label{fig:blackbox_inside}
\end{figure}

To ensure compliance with the time constraint, $c_c(\theta^i)$ in the event of a time constraint violation must be higher than any possible strain-related cost $\alpha_d c_s(\theta^i)$. Therefore, $\alpha_d$ is set as the inverse of the difference between the maximum and minimum strain values encountered in the \textit{whole} dataset $\mathcal{D}$. This ensures that $\alpha_d c_s(\theta^i) \leq 1$ in all cases while $c_c(\theta^i) \geq 1$ if $t^i_{\mathrm{st}} \geq \bar{t}_\mathrm{st} $. \modif{The different components of the black box are illustrated in Figure \ref{fig:blackbox_inside}.}

\subsection{Active learning and optimization}
The black-box approach described in Section~\ref{sec:virtual_sensor} assumes the existence of a dataset $\mathcal{D}$ of measured trajectories $\{\tau_M\}$. However, in the active learning problem described in Section~\ref{sec:problem_description}, no preliminary data are provided. Instead, an instrumented turbine is available, making it possible to incrementally construct the dataset with a budget of $N_{\mathrm{b}}$ startup trajectories. The dataset $\mathcal{D}^j$ is defined as the set containing measured trajectories $\{ \tau^0_M, \tau^1_M, ..., \tau^j_M \}$, and the $m$ model parameters $\phi^j$ as those trained over $\mathcal{D}^j$.

\begin{figure}
    \centering
    \includegraphics[width=0.8\linewidth]{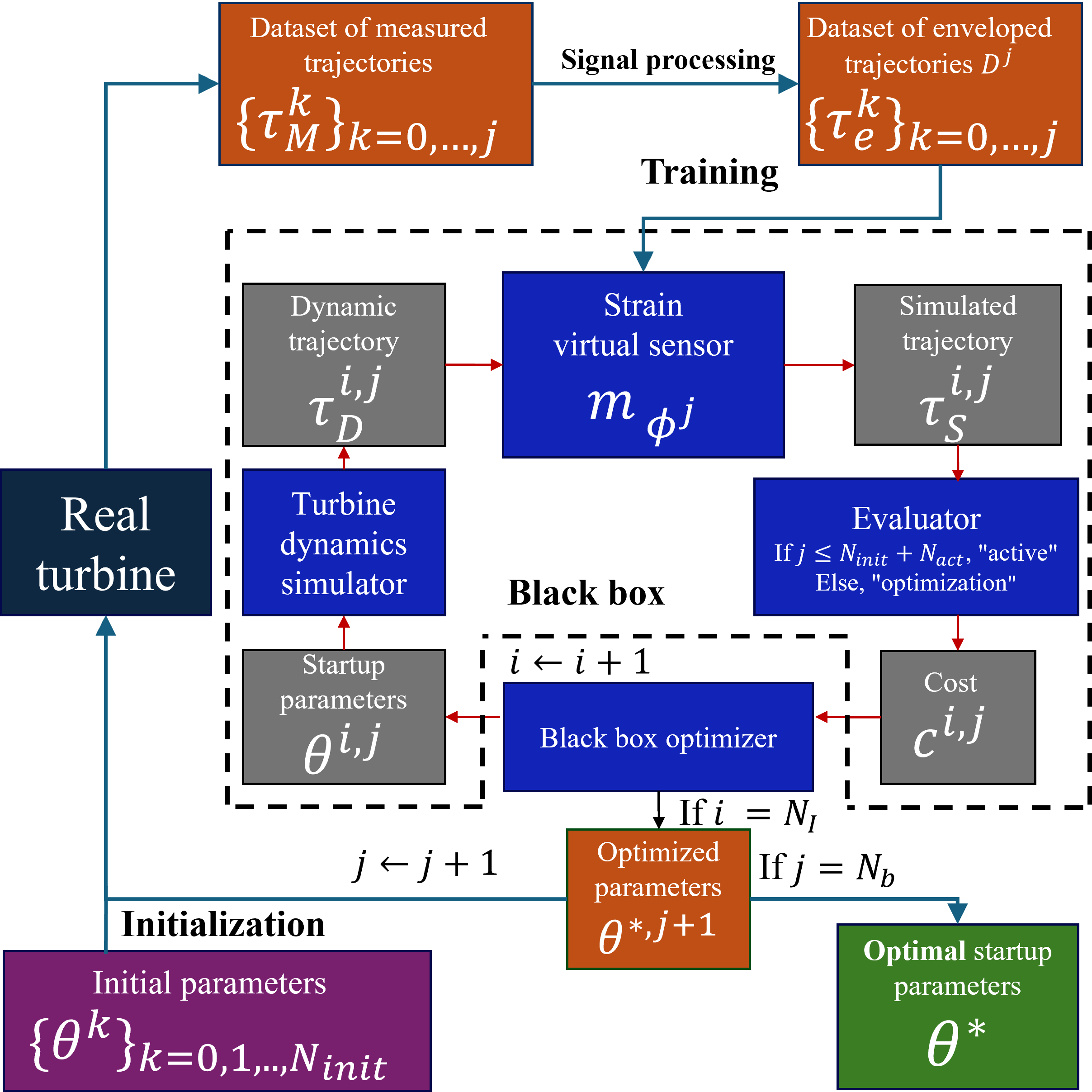}
    \caption{Active learning nested loop for startup parameter optimization.}
    \label{fig:active_optimization_nestedloop}
\end{figure}

The proposed active scheme is presented in Figure~\ref{fig:active_optimization_nestedloop}, where the black box optimizer is used at every outer loop iteration to provide the parameters $\theta^{j+1}$ to test on the turbine. The new measured trajectory is then added to the dataset, and the virtual sensor is retrained with the new data. The process goes through three phases with predetermined trajectory budgets: initialization ($N_{init}$), active learning ($N_{act}$), and optimization ($N_{opt}$), such that $N_{init} + N_{act} + N_{opt} = N_{\mathrm{b}}$. For the remainder of the paper, the variables indexed by $i$ in the single black-box loop will be indexed by ${i,j}$ to include nested loop index $j$.

To kick-start the process, the turbine is run on $N_{init}$ trajectories using pre-selected parameters $\{\theta^k\}_{k = 0, ..., N_{init}}$. These parameters are chosen to provide relevant data to the virtual sensor. Ideally, they generate a fast startup with high opening values and a slow startup with low opening values to map the extremes of the trajectory space. It can also be useful to include another trajectory in between, such as a standard startup with the usual parameters used on the turbine.

The objective of the active learning phase is to explore the space of trajectories in order to supply the most relevant data for training the virtual sensor. To do so, the same black-box optimization system as presented in Section~\ref{sec:black-box_simple} is used. However, the evaluator is modified to account for the epistemic uncertainty, expressed as the standard deviation $\sigma_{\mathrm{ep}, n}^{\{u,l\},i,j}$ of the ensemble's individual $m$ model predictions of sum $\hat{\mu}^{\{u,l\},i,j}_n + \hat{\sigma}^{\{u,l\},i,j}_n $. Inspired by the upper confidence bound approach \citep{Auer_2002} for exploration, the strain cost in Equation~(\ref{eq:strain_cost}) is modified as follows: 

\begin{equation}
        c_{s, act}(\theta^{i,j}) = \max_{n \in \nu_S} \left(\hat{\mu}^{u,i,j}_n + \hat{\sigma}^{u,i,j}_n - 2\sigma_{\mathrm{ep}, n}^{u,i,j} \right) - \min_{n \in \nu_S}  \left(\hat{\mu}^{l,i,j}_n + \hat{\sigma}^{l,i,j}_n + 2\sigma_{\mathrm{ep}, n}^{l,i,j} \right)
\end{equation}

By reducing the envelope by twice the standard deviation of the epistemic uncertainty, the evaluator assumes the best-case scenario in light of the model's uncertainty. This optimism in the face of uncertainty encourages promising trajectories that combine low estimated strains and strong potential for improvement.

Once the virtual sensor has been fed with enough data to be trusted in the relevant regions of the input space, the third phase seeks to obtain optimal parameters $\theta^*$ for the turbine. In this case, the model's epistemic uncertainty is not considered. Due to the inherent stochastic nature of the strain process, one or two loops can be added to the process by incrementing the dataset with the newly obtained data to ensure convergence. The last iteration is used to test the performance of the final parameters on the turbine.

\section{Experimental results}
\label{sec:experimental_results}

\subsection{Experimental setup}
Our case study is an instrumented Francis turbine runner installed in a storage hydroelectric power plant. The specific speed of the turbine is $n_{QE} = n Q^{0.5} / (gH)^{3/4} = 65$, with $n$, the runner frequency, $Q$, the nominal flow rate, $g$, the gravity and $H = 72$ m, the nominal head. For this power plant, the time limit for reaching synchronous speed was set to $\bar{t}_\mathrm{st} = 90$s. 

\begin{figure}[h]
    \centering
    \includegraphics[width=0.9\textwidth]{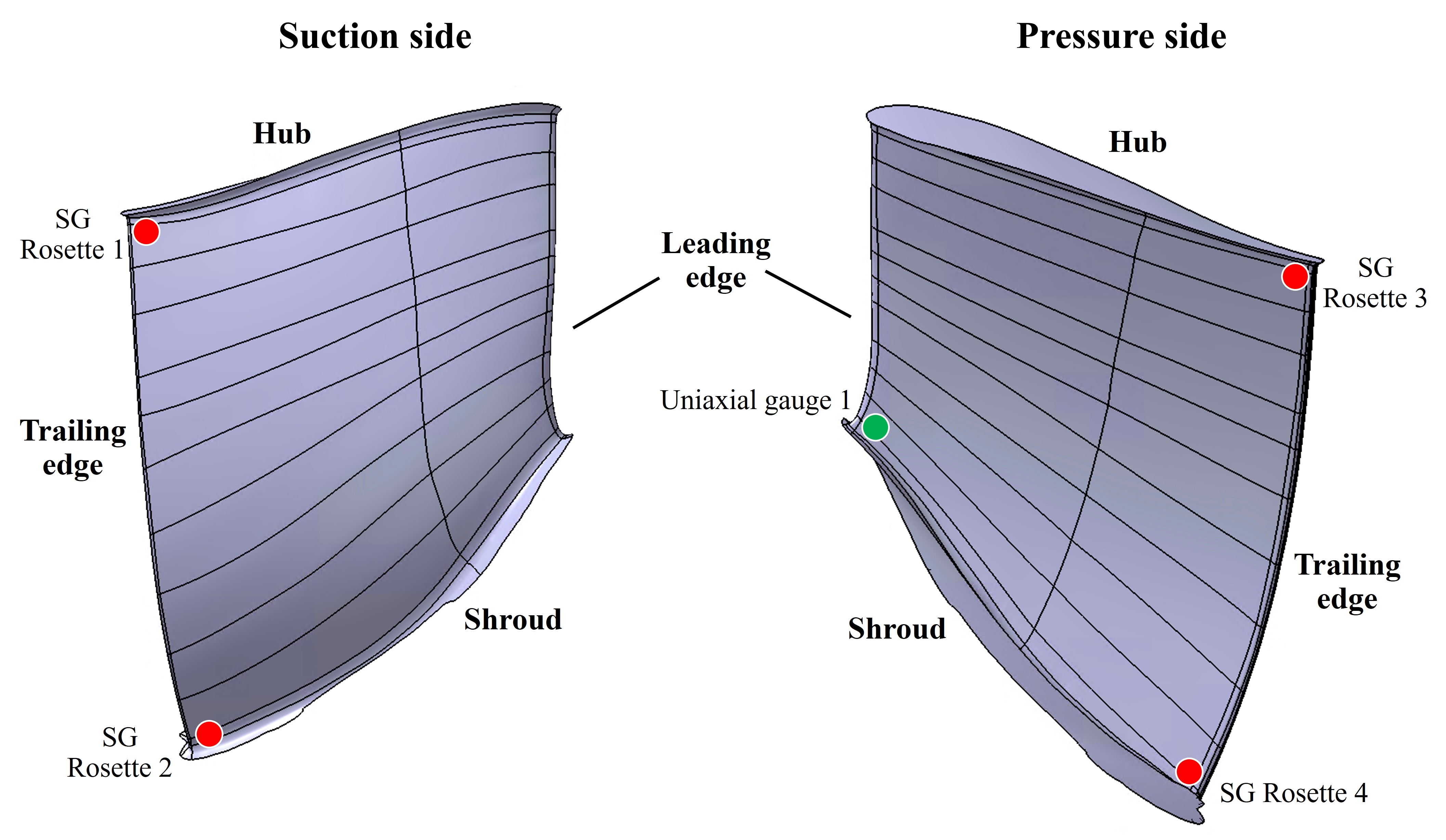}
    \caption{Positions of the uniaxial gauge and rosettes on the Francis turbine runner, as indicated by green and red dots.}
    \label{fig:position_jauge}
\end{figure}

Two blades were instrumented with four strain gauge rosettes near the trailing edge, with two rosettes positioned on the suction side and two on the pressure side. An additional uniaxial strain gauge was also placed on the pressure side near the leading edge. The positions of the uniaxial gauge and the rosettes are illustrated in Figure~\ref{fig:position_jauge}. \modif{The typical setup for data acquisition for such measurement campaign, where sensor data on the rotating part can be transmitted either via slip-ring or through a Wi-Fi telemetry system for on-site processing, can be found in \citep{marcouiller_obtaining_2015}.}

Due to blade geometry and the presence of fillets, the gauges could not be placed directly at the stress hotspots, where the strain amplitudes are largest; instead, they were positioned at the nearest locations permitted by the geometrical constraints. \modif{The identification of stress hotspots location was based on CFD and FEA simulations using state-of-the-art numerical procedures as described in \citep{Doujak2022}. In the turbine used as test-case, two static stress hotspots were identified near the trailing edge of the blades, at the fillets between the shroud / hub and the blades, which are typical hotspot position for Francis turbines.}     

The turbine guide vanes were equipped with a linear encoder to measure their opening angle $o$, while the turbine rotational speed $\omega$ was measured by a proximeter. 

\begin{figure}[ht]
    \centering
    \includegraphics[width=0.72\textwidth]{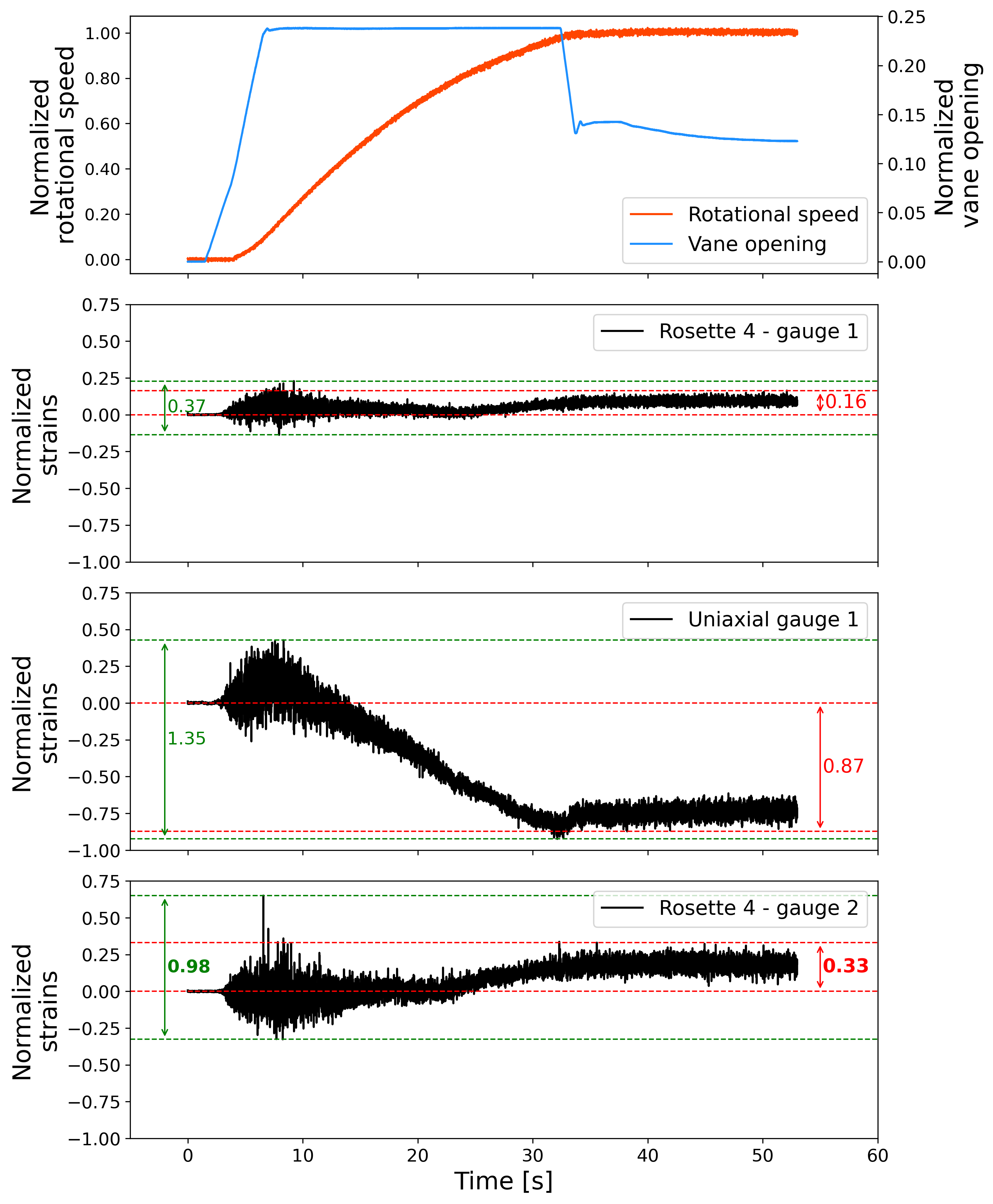}
    \caption{Standard startup trajectory measurements, including the strains for three different gauges. The largest cycle during startup is shown between the green lines. The irreducible cycle between standstill and speed-no-load is shown in red. Rosette 4 -- gauge 1 is an example of low potential because of the strain amplitude. The potential of uniaxial gauge 1 is low because of the high strain difference between standstill and speed-no-load. Rosette 4 -- gauge 2 shows the highest potential and was chosen for the optimization.}
    \label{fig:rosette}
\end{figure}

The optimization is conducted using strain data measured by gauge 2 of rosette 4, located on the pressure side, near the trailing edge and shroud. This sensor measured dynamic strains with the highest potential for optimization, as shown in Figure~\ref{fig:rosette} with strain signals from three different gauges. That potential was determined based on the highest strain cycle during startup, along with the irreducible strain cycle between standstill and the moment when the turbine is ready to be synchronized (speed-no-load). 

Other experiments conducted during the measurement campaign allowed us to test $N_{init} = 5$ startups for initial trajectories, which included two standard startups and three alternatives designed to cover the space of possible trajectories. The optimization process was run using a budget of $N_{act} = 2$ active learning startups and $N_{opt} = 1$ optimization trajectory. A last optimization trajectory was run with $\bar{t}_\mathrm{st} = 60$ s to study the effect of shorter time constraints. 

The speed governor has predefined limits for every startup parameter, which were provided to the black-box optimizer. Additional limits were set on $o_\mathrm{trigger}$ and $\omega_\mathrm{trigger}$ to prevent turbine overspeed; these were not enforced for the initial startups. The parameter limits for optimization are shown in Table~\ref{tab:limit_parameters}. 

\begin{table}[]
    \centering
    \begin{tabular}{|c|c|c|}
    \hline
       Parameters  & Min & Max \\
       \hline
       $r_o$ (\%/s) & 1 & 10 \\
       $o_\mathrm{ini}$ & 0 & 0.34 \\
       $\omega_\mathrm{trigger}$ & 0 & 0.95 \\
       $o_\mathrm{trigger}$ & 0 & 0.21 \\
       \hline
    \end{tabular}
    \caption{Startup Parameter Limits for Black-Box Optimization}
    \label{tab:limit_parameters}
\end{table}

The black-box optimization loop was given a budget of $N_I = 200$ iterations. NOMAD was launched on 10 parallel threads using the PSD-MADS algorithm \citep{Audet_Dennis_Digabel_2008}. The various trajectory sampling frequencies were set to balance accurate modeling and time efficiency. Trajectories $\tau_M$ were measured at 5,000 Hz and then resampled at $f_M =$ 500 Hz. The enveloped trajectories $\tau_e$ were calculated with a window $w$ corresponding to 10 s, and then resampled at $f_e$ = 10 Hz. On the black-box side, the turbine dynamics simulator created dynamic trajectories $\tau_D$ at $f_D = 10$ Hz, with $f_S = f_D$ for simulated trajectories $\tau_S$.


\subsection{Results}

\begin{figure}
     \centering
        \includegraphics[width=1\linewidth]{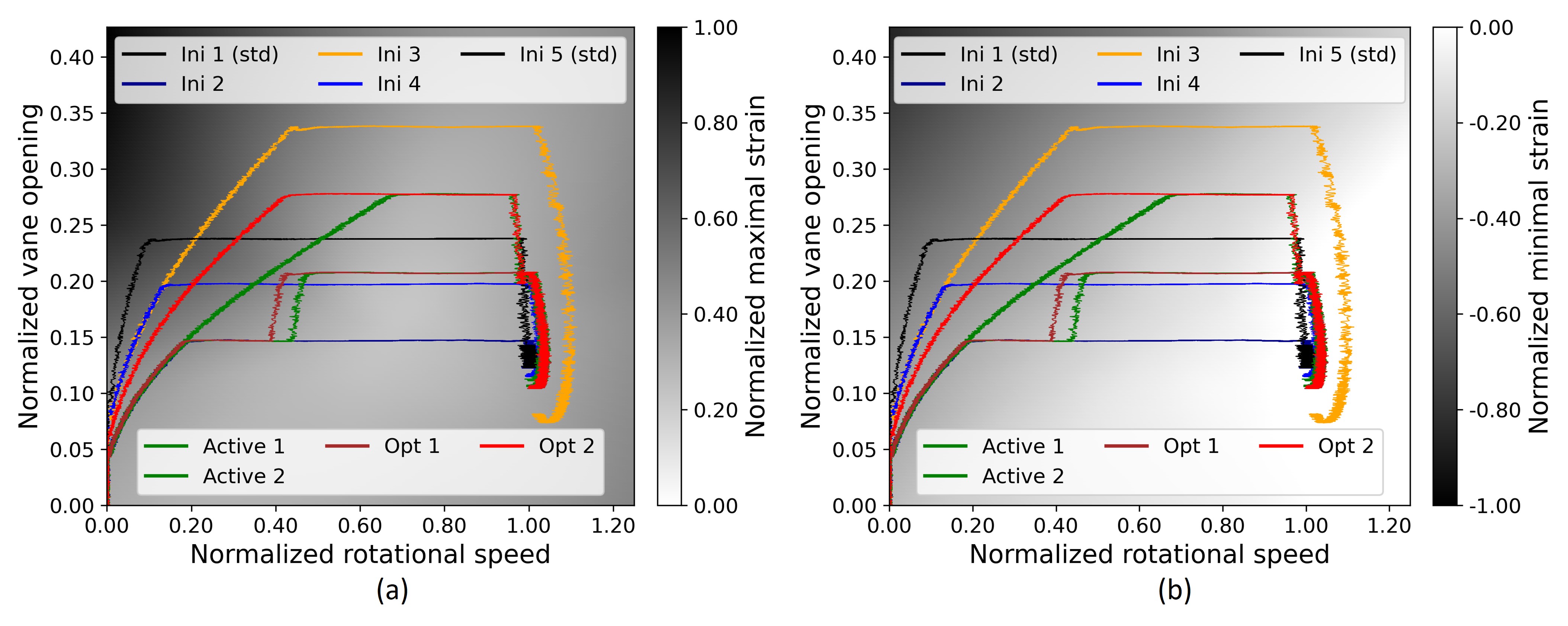}
        \caption{\modif{\modif{Measured} turbine dynamic trajectories mapped on the opening - rotational speed space. The color map (white is better) represents the latest trained virtual sensor's predictions of $\hat{\mu} + \hat{\sigma}$ for the upper (a) and lower bounds (b) of the strain envelope.}}
        \label{fig:results_map}
\end{figure}

The sequential startup parameters $\theta^j$ are given in Table~\ref{tab:startup_parameters}, together with the largest observed strain cycle. Figure~\ref{fig:results_map} provides the trajectories projected on the $(o, \omega)$ space. The color map shows the latest trained virtual sensor's predictions of $\hat{\mu} + \hat{\sigma}$ for the upper (\modif{Figure~\ref{fig:results_map}a}) and lower (\modif{Figure~\ref{fig:results_map}b}) bounds of the strain envelope. \modif{The highest amplitudes of the strain envelope are observed in operating zone with low rotational speed and higher vane opening, similar to the results presented by Gagnon et al. \citep{Gagnon_Pham_Mai_Favrel_2023}. This behavior may be explained by the increased stochasticity of the flow in this operating regime, originating from the shedding and collapse of vortices as proposed by Morissette and Nicolle \citep{Morissette2019}. However, a more comprehensive understanding of this correlation would require an in-depth flow analysis through unsteady CFD-FEA of different startup scenarios, which is beyond the scope of the present paper.} 

\begin{figure}
    \centering
    \includegraphics[width=\linewidth]{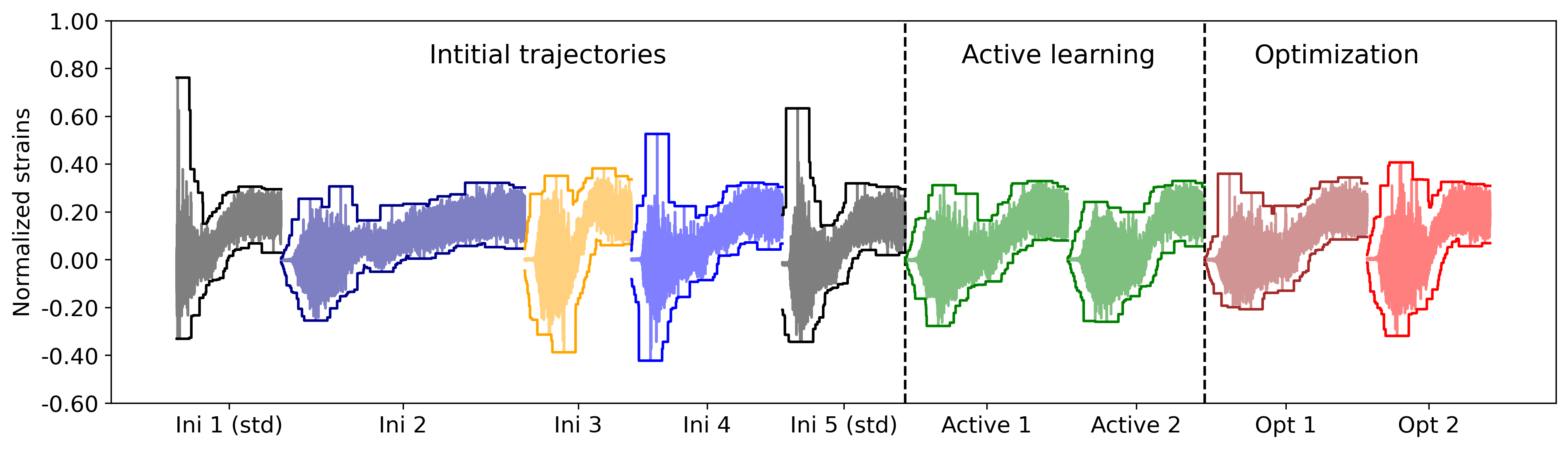}
    \caption{Strain measurements $\tau_M$ and envelopes $\tau_e$ for the sequential startups run on the turbine. The optimized startups have significantly lower strain amplitude than the standard ones.}
    \label{fig:results_strain}
\end{figure}

Figure~\ref{fig:results_strain} shows the optimization process's measured strains on the various startups over time.

\begin{table}[]
    \centering
    \begin{tabular}{|c|c c c c |c c|}
       \hline
       Startup  & $r_o$ & $o_{\mathrm{ini}}$  & $\omega_{\mathrm{trigger}}$ & $o_{\mathrm{trigger}}$ & $t_\mathrm{st}$ & Largest cycle \\
       \hline
       Initial 1 (std.) & 10 & 0.24 & 0.97 & 0.15 & N/A & 1.09\\
       Initial 2 & 1 & 0.15 & 0.8 & 0.15 & 105 s & 0.58 \\
       Initial 3 & 2.5 & 0.34 & 0.8 & 0.34 & 46 s & 0.77\\
       Initial 4 & 2.5 & 0.20 & 0.8 & 0.20 & 65 s & 0.95
       \\
       Initial 5 (std.) & 10 & 0.24 & 0.97 & 0.15 & 53 s & 0.98\\
       \hline
       Active 1 & 1 & 0.15 & 0.42 & 0.21 & 70 s & 0.61 \\
       Active 2 & 1 & 0.28 & 0.95 & 0.21 & 59 s & 0.59 \\
       \hline 
       Optimal 1 (90 s) & 1 & 0.15 & 0.37 & 0.21 & 70 s & 0.57\\
       Optimal 2 (60 s) & 1.69 & 0.28 & 0.95 & 0.21 & 53 s & 0.73\\
       \hline
    \end{tabular}
    \caption{Parameters $\theta$ and Results of the Startup Optimization Sequence}
    \label{tab:startup_parameters}
\end{table}

The standard parameters (Initial 1 and 5) results in a large strain amplitude as they pass through the high-opening, low-rotational speed region. An issue during Initial 1 delayed the recording of measurements, which were lost for the first few seconds of the startup. The largest cycle difference between the two startups with standard parameters is a result of the process's stochasticity. Initial 2 follows the lowest limit for each of the opening parameters. It passes through a low-strain area, but is too slow to respect the time constraint. Initial 3 instead explores the opening values' upper limit, with a relatively high opening rate.

\begin{figure}
     \centering
    \includegraphics[width=0.9\linewidth]{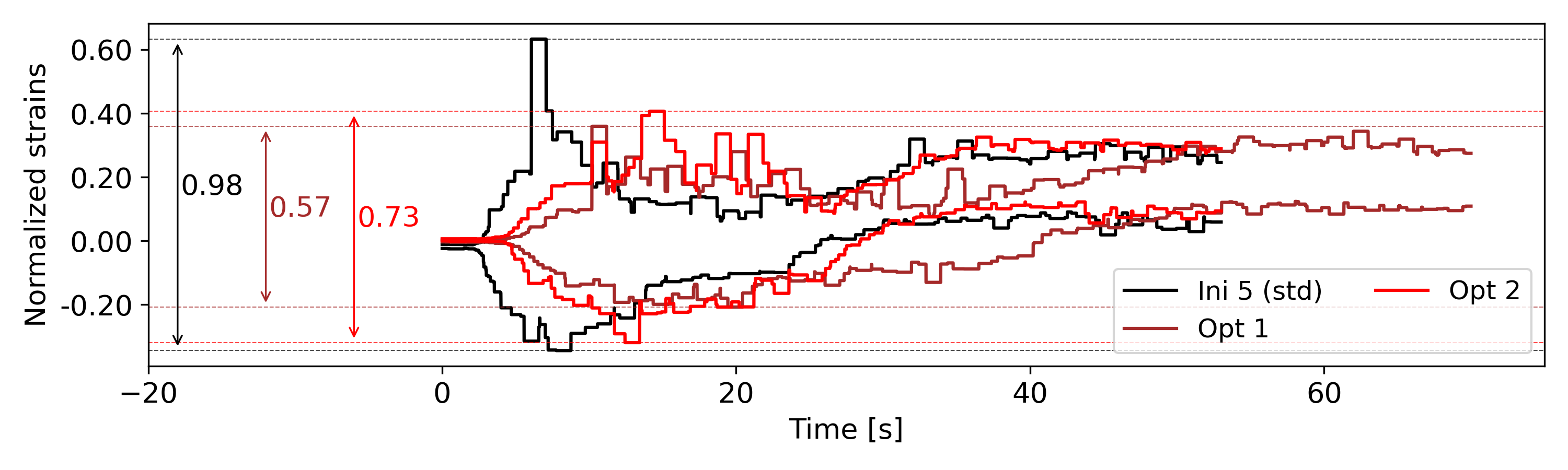}
        \caption{\modif{Comparison of measured strain envelopes for initial (\textit{Ini 5}) and optimized startups (\textit{Opt 1} is limited to $\bar{t}_{\mathrm{st}} = 90$s, \textit{Opt 2} is limited to $\bar{t}_{\mathrm{st}} = 60$s). While the optimization process is also able to reduce the strain amplitude by 26\% for similar durations to the initial startup (\textit{Opt 2}), it is also able to take advantage of a longer startup time to improve the performance to 42\% (\textit{Opt 1}).}}
        \label{fig:results_details}
\end{figure}

The active learning trajectories already show a significant strain reduction. Clearly, they follow the lowest opening rate of Initial 2, and catch up for time afterwards. Interestingly, the startup duration $t_\mathrm{st}$ is always shorter than the time constraint. This could be due to the small penalty added to time as seen in Equation~(\ref{eq:time_cost}). 
The Optimal 1 startup is even better, leading to a 42\% reduction compared to the best standard startup. Finally, Optimal 2 contends with the need to open the vanes more widely to reach the 60 s time constraint. It still attempts to keep $r_o$ lower than most initial trajectories, preferring instead to increase $o_\mathrm{init}$ to catch up in time. Optimal 2 still achieves a 26\% reduction in strain amplitude compared to the standard parameters, with equal time to reach synchronous speed. \modif{The strains measured during the optimal trajectories are shown in higher temporal resolution in figure \ref{fig:results_details}.}

\begin{figure}
    \centering
    \includegraphics[width=0.9\linewidth]{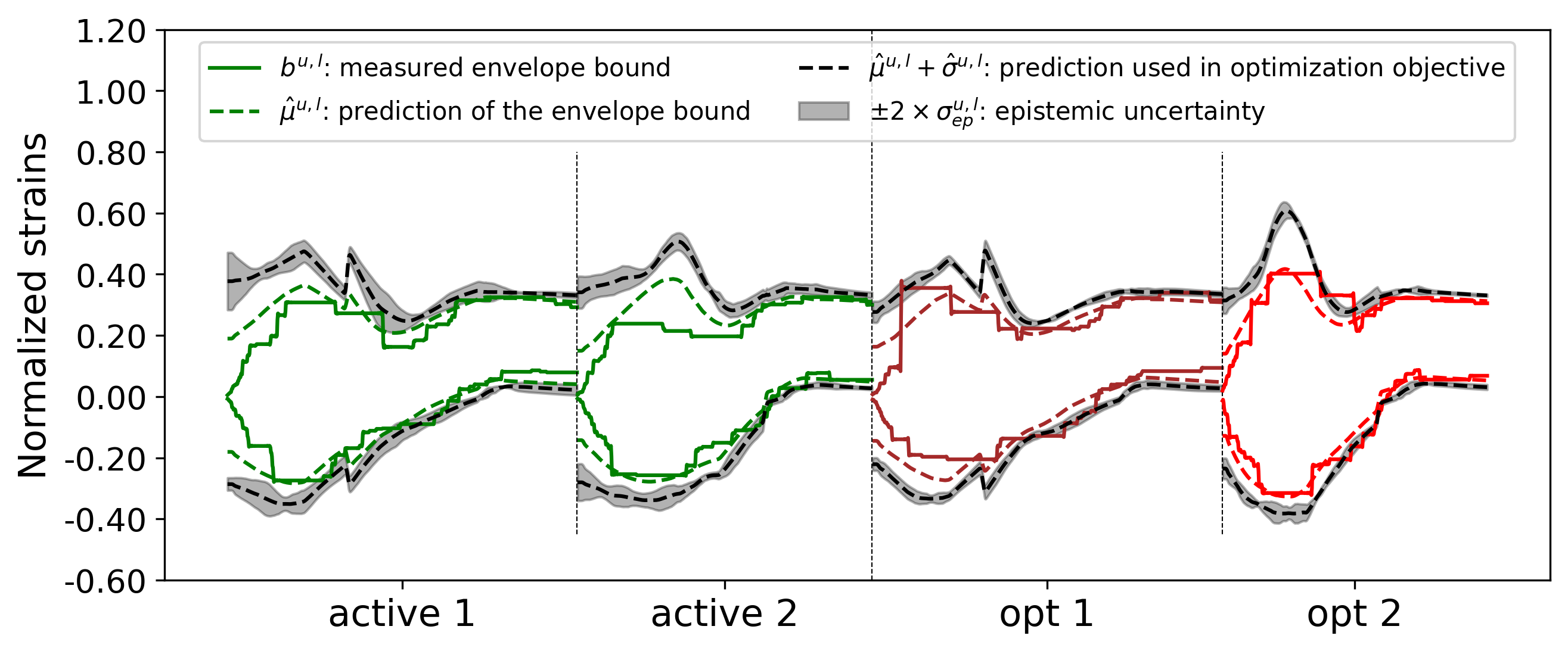}
    \caption{Black box strain predictions given optimal startup parameters $\theta^{*, j}$ compared to measured strain envelopes.}
    \label{fig:results_virtual_sensors}
\end{figure}

Figure~\ref{fig:results_virtual_sensors} presents the strain envelope predictions of the optimal trajectory $\tau^{*,j}_S$ modeled in the black box corresponding to the optimized startup parameters $\theta^{*,j}$ of the active learning and optimization phases, and compares them to the measured envelope bounds $b^{u,l}$. For each startup $j$, the strain prediction is made by model $m_\phi^{j-1}$ trained on the previous data, as per the active optimization process.  While the mean prediction $\hat{\mu}^{u,l}$ of the envelope is generally accurate, the stochasticity of the process induces an significant standard deviation $\hat{\sigma}^{u,l}$. Using $\hat{\mu}^{u,l} + \hat{\sigma}^{u,l}$ in the optimization objective $c_s(\theta^{i,j})$ ensures that the expected maximum strain cycle is not underestimated.

\paragraph{Running time}
The optimization process was run on a laptop computer with an Intel i7-11850H 2.5GHz CPU, 64GB of RAM and an NVIDIA RTX A2000 GPU. One external optimization loop, from the reception of the new measurements to the output of the optimized parameters, lasted on average 9~minutes, which is acceptable in a time-constrained context such as a measurement campaign. \\


\section{Discussion}

The results show that the proposed active learning and optimization method is promising and could be applied systematically as part of the commissioning of a new turbine. \modif{An interesting aspect of the method is that the learned model, which can be subject to accuracy errors and is usually delicate to implement in critical industrial settings, is only used to guide the optimization process. The actual strain is measured experimentally on the turbine, guaranteeing the performance of the optimized parameters.} However, several questions remain to be studied.  First, the relevance of the active learning phase depends on the optimization landscape and the number of initial trajectories. In the presented use case, the landscape was smooth and the initial trajectories covered a good portion of the space. The epistemic uncertainty values were low, and the active learning phase, including the results, was not significantly different from the optimization phase. Second, our approach was applied to the strain amplitude measured at a single strain gauge. It is possible that minimizing the strain at one location could increase the strain at another location. A sounder approach would perform optimization over all the installed gauges. The challenges in achieving such a generalized optimization are the computation requirements and robustness to potential defects occurring in strain gauges during the optimization campaign.
Third, the water head may influence the turbine strain during startup. During a measurement campaign taking place over a few days, this condition can remain quite constant, but this is not the case during turbine operation. We hypothesize that parameters leading to lower strains for a given water head generalize to other conditions, which remains to be confirmed.


\section{Conclusion}
\label{sec:conclusion}

In this paper, we have proposed an automated approach to optimizing hydroelectric generating units (HGU) startup sequences aimed at minimizing stresses by using a limited budget of measured startup sequences during on-site prototype measurement. The approach combines active learning and black-box optimization techniques, utilizing virtual strain sensors and dynamic HGU simulations.

The methodology was validated in real time by means of an on-site measurement campaign on an instrumented Francis turbine prototype. The results demonstrated the efficacy of our active learning approach, as the parameters identified for optimal startup reduced the highest measured strain cycle amplitude by 42\%, reducing the fatigue damage incurred by the turbine during startup.

This method shows great promise for systematic application during the commissioning of new turbines, potentially contributing to the extension of their operational lifespans. This is particularly relevant in the context of the increasing number of startup sequences necessitated by the integration of intermittent renewable energy sources.

However, our approach has several limitations, which include optimizing only on a single strain gauge and ignoring the potential impact of water head and downstream reservoir level. These limitations must be addressed in future studies, potentially through reduced-scale model testing and additional prototype measurements campaigns.

Overall, our work paves the way for the development of more efficient HGU startup optimization strategies, thereby promoting the role of hydropower as a key contributor to the integration of intermittent renewable energy sources.

\bibliographystyle{ieeetr} 
\setcitestyle{square}
\bibliography{main.bib}


\newpage
\appendix
\setcounter{table}{0}

\section{Notation table}
\label{app:notation_table}
\begin{table}[h]
    \centering
    \begin{tabular}{|c|c|c|}
    \hline
    \multirow{16}{*}{\centering HGU dynamics}  & $n$ & Time step \\
    & $\omega_n$ & Turbine rotation speed \\
     & $o_n$ & Guide vanes opening \\
     & $u_n$ & Guide vanes opening setpoint \\
     & $\theta$ & Startup parameters \\
    & $r_o$ & Opening rate \\
    & $o_\mathrm{ini}$ & Initial opening \\
    & $\omega_\mathrm{trigger}$ & Trigger rotation speed \\
    & $o_\mathrm{trigger}$ & Trigger opening \\
     & $\omega_S$ & Synchronous rotation speed \\
    & $t_\mathrm{st}$ & Time to reach $\omega_S$ \\
    & $\bar{t}_\mathrm{st}$ & Maximum time to reach $\omega_S$ \\
    & $f, g, h$ & Dynamic functions of $o, \omega$ and $u$ \\
    & \modif{$T_{turb}$} & \modif{Turbine torque} \\
    & \modif{$J$} & \modif{Inertia of rotating components} \\
    & \modif{$q_{4\cdots N_{eq}}$ }& \modif{Internal speed governor model variables} \\
    \hline
    \multirow{3}{*}{\centering Strains}  & $s_n$ & Strain at $n$ \\
    & $s^{\{u, l\}}_n$ & Upper/lower strain envelope bound \\
    & $b^{\{u,l\}}$ & Probability density functions of $s^{\{u, l\}}$ \\
    \hline
    \multirow{7}{*}{\centering Trajectories}  & $\tau_D$ & Dynamic trajectory ($\omega_n$ and $o_n$) \\
    & $\tau_M$ & Measured trajectory ($\omega_n, o_n$ and $s_n$) \\
    & $\tau_e$ & Enveloped trajectory ($\omega_n, o_n, s^u_n, s^l_n$) \\
    & $\tau_S$ & Simulated trajectory ($\omega_n, o_n, \hat{\mu}^u_n, \hat{\sigma}^u_n, \hat{\mu}^l_n, \hat{\sigma}^l_n,$)\\
    & $f_{\{D,M,e,S\}}$ & Trajectory sampling frequency \\
    & $n_{\{D,M,e,S\}}$ & \# time steps in trajectory \\
    & $\nu_{\{D,M,e,S\}}$ & Set of time steps (0, ..., $n_{\{D,M,e,S\}}$) \\ 
    \hline
    \end{tabular}
    \caption{Notation table -- Part I}
    \label{tab:notation}
\end{table}

\begin{table}[h]
    \centering
    \begin{tabular}{|c|c|c|}
    \hline
    \multirow{6}{*}{\centering Strain model}  & $\mathcal{D}$ & Dataset of enveloped trajectories $\{\tau_e\}$ \\
    & w & Envelope window \\
    &$m_\phi$ & Neural network model with param. $\phi$ \\
    & $\hat{\mu}^{\{u,l\}}$ & Mean prediction \\
    & $\hat{\sigma}^{\{u,l\}}$ & Aleatoric standard deviation prediction \\
    & $\sigma_{\mathrm{ep}}^{\{u,l\}}$ & Epistemic ensemble standard deviation \\
    \hline
    \multirow{10}{*}{\centering Optimization}  & $i$ & Inner loop index (simulated traj.) \\
    & $j$ & Outer loop index (measured traj.) \\
    & $*$ & Optimal index \\
    & \modif{$\mathcal{L}$} & \modif{Objective function for strain} \\
    & $c(\theta)$ & Cost function for the optimizer \\
    & $c_s(\theta)$ & Cost due to strain \\
    & $c_{s,\mathrm{act}}(\theta)$ & Active learning variant of $c_s(\theta)$ \\
    & $c_c(\theta)$ & Cost due to time constraint \\
    & $\alpha_d$ & Cost function normalization factor \\
    & $N_{\{b, ini, act, opt\}}$ & Measured traj. (outer loop) budget \\
    & $N_I$ & Simulated traj. (inner loop) budget\\
    \hline
    \end{tabular}
    \caption{Notation table -- Part II}
    \label{tab:notation2}
\end{table}

\end{document}